\documentclass[sigconf,nonacm]{acmart}
\usepackage[utf8]{inputenc}

\AtBeginDocument{%
  \providecommand\BibTeX{{%
    \normalfont B\kern-0.5em{\scshape i\kern-0.25em b}\kern-0.8em\TeX}}}

\setcopyright{acmlicensed}
\copyrightyear{2022}
\acmYear{2022}
\acmDOI{10.1145/3522664.3528608}

\acmConference[CAIN'22]{1st Conference on AI Engineering - Software Engineering for AI}{May 16--24, 2022}{Pittsburgh, PA, USA}
\acmPrice{15.00}
\acmISBN{978-1-4503-9275-4/22/05}

\begin{document}

\title{Practical Insights of Repairing Model Problems on Image Classification}

\author{Akihito Yoshii}
\email{{yoshii.akihito, tokumoto.susumu}@fujitsu.com}
\author{Susumu Tokumoto}
\affiliation{%
  \institution{Fujitsu Limited}
  \city{Kawasaki}
  \state{Kanagawa}
  \country{Japan}
  \postcode{43017-6221}
}

\author{Fuyuki Ishikawa}
\affiliation{%
  \institution{National Institute of Informatics}
  \city{Tokyo}
  \country{Japan}}
\email{f-ishikawa@nii.ac.jp}

\begin{abstract}
  Additional training of a deep learning model can cause negative effects on the results, turning an initially positive sample into a negative one (degradation). Such degradation is possible in real-world use cases due to the diversity of sample characteristics. That is, a set of samples is a mixture of critical ones which should not be missed and less important ones. Therefore, we cannot understand the performance by accuracy alone. While existing research aims to prevent a model degradation, insights into the related methods are needed to grasp their benefits and limitations. In this talk, we will present implications derived from a comparison of methods for reducing degradation. Especially, we formulated use cases for industrial settings in terms of arrangements of a data set. The results imply that a practitioner should care about better method continuously considering dataset availability and life cycle of an AI system because of a trade-off between accuracy and preventing degradation.
\end{abstract}



\maketitle

\section{Degradation in Deep Learning Tasks}
A deep learning model can be trained multiple time (retrained) with additional datasets. Considering a classification task, the model accuracy is expected to be increased after the retraining.

However, a part of samples can change from a positive (classified correctly) to a negative (classified incorrectly) even if the accuracy increases totally. We call such a situation as "degradation" in this talk. Especially in a real-world setting, available additional datasets are varying in size, quality and importance; and thus, a model can be susceptible to a degradation-causing sample.

In a use case that a misclassification of samples leads to serious accident, those samples should be maintained to be positive. However, such situation cannot always be measured with accuracy. This is because even with a highly accurate model which classifies almost all samples correctly, the model is not acceptable if the most of “important” samples are belong to the misclassified ones. 

\section{Existing Approaches and Our Motivation}
In order to reduce the degradation, some works improve retraining procedure whereas others propose different approaches. Our existing work, NeuRecover\cite{tokui2022neurecover} is a non-retraining method which enables a developer to repair a part of a DNN model weights with smaller data samples. On the other hand, crafting retraining is another approach to reduce degradation. For example, Backward Compatibility ML\cite{bansal2019updates}\cite{srivastava2020an} introduces a penalty term to a loss function. It is designed to prevent new errors which did not exist before a retraining which affects a user expectation\cite{bansal2019updates}.

We need to grasp advantages and limitations of different methods because they have similar characteristics considering the reduction of the degradation while they are developed by different focuses.

In this talk, we will present implications on suitable use cases for each method, derived from a comparison of methods for reducing degradation. This discussion is important because maintaining model quality leads to a reliable AI-powered software performance. Although an accuracy is one of metrics indicating model performance, a trade-off exists between an accuracy and certain kinds of others. Hypothesizing real use cases will be a clue for making decision under the trade-off and achieving an intended task.

\begin{figure*}[h]
    \centering
    \includegraphics[scale=0.45]{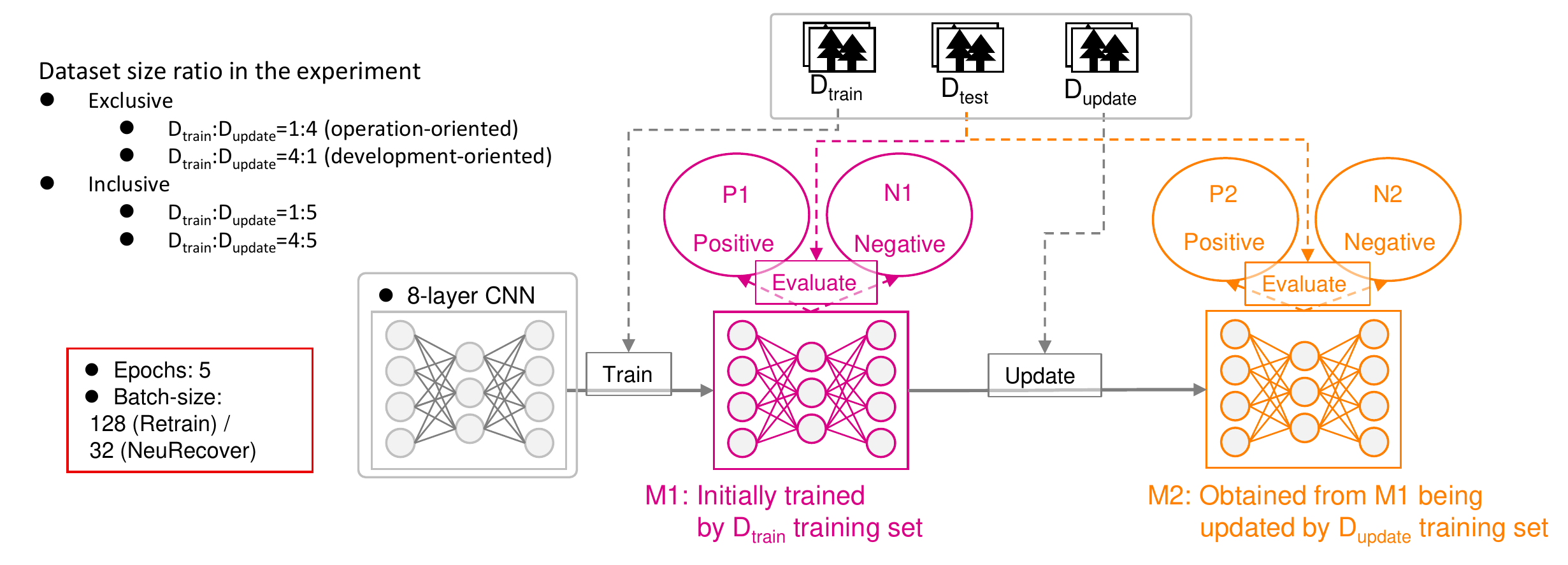}
    \caption{Overall processes}
    \label{fig_oap}
    \Description{Overall processes}
\end{figure*}

\section{Experiment and Results}
We formulated three use cases from an aspect of dataset arrangement: exclusive (operation-oriented), exclusive (development-oriented) and inclusive.

In an operation-oriented use case, an update set ($D_{update}$) is larger than a training set ($D_{train}$) while a development-oriented case is the opposite. The "Exclusive" assumes no intersection between $D_{train}$ and $D_{update}$; on the other hand, $D_{train}$ is included by larger $D_{update}$ under the "Inclusive" condition.

Figure \ref{fig_oap} shows an overall process of the experiment. We executed two retraining methods (categorical cross entropy and Backward Compatibility\footnote{The loss function was imported from https://github.com/microsoft/BackwardCompatibilityML whose version is 1.4.2, with the reduction parameter was set to SUM\_OVER\_BATCH\_SIZE.} ML\cite{bansal2019updates}\cite{srivastava2020an}) and one non-retraining method (NeuRecover\cite{tokui2022neurecover}) with datasets arranged based on the aforementioned conditions. We used CIFAR10\cite{Krizhevsky09learningmultiple} and Fashion MNIST\cite{DBLP:journals/corr/abs-1708-07747}; and then split them into $D_{train}$ and $D_{update}$.

We evaluated results with a test set apart from $D_{train}$ and $D_{update}$ in each dataset and metrics proposed in related work. Especially we compared BR (Break Rate)\cite{tokui2022neurecover}, BEC (Backward Error Compatibility)\cite{srivastava2020an} and accuracy improvement rate. BR is defined as $\frac{n(P_1\cap N_2)}{n(P_1)}$\cite{tokui2022neurecover} and BEC is defined as $\frac{n(N_1\cap N_2)}{n(N_2)}$\cite{srivastava2020an}. Accuracy improvement rate can be calculated by $\frac{(M1 accuracy)}{(M2 accuracy)}$.

The results suggest that NeuRecover has relatively high BEC and BR while Accuracy improvement rate is lower than others. The difference of Accuracy improvement rate among methods is smaller under the $D_{train}:D_{update}=4:1$ condition than $D_{train}:D_{update}=1:4$ condition. On the other hand, BC(ne) keeps stably higher even when BR is higher and the accuracy improvement is small.

\section{Lessons Learned}
We observed following lessons from the results of our experiment.
    \begin{description}
        \item[Methods depending on a purpose and a dataset arrangement]{
        Suitable method can be chosen in an aspect of metrics stated in the previous section and each experiment condition.
        
        A possible scenario of the "exclusive" situation is that only a pre-trained model is available for a developer and the pre-trained model is updated with a dataset prepared by the developer at the development / operation time. On the other hand, an "inclusive" situation assumes that the developer has a dataset available from the beginning without limitation; therefore the developer can include initial training data in a dataset for future updates.
        
If an environment change at an operation time are so significant that a model need to be retrained, "exclusive (operation-oriented)" case can be assumed. While an "exclusive (development-oriented)" is for the situations that an environment is relatively static.}
        \item[Accuracy does not always earn the highest priority]{A practitioner should cope with a trade-off between accuracy and preventing degradation. For example, Bansal et al. are pointing out the discrepancy between human expectation of AI system and model updates \cite{bansal2019updates}. Thus, higher accuracy and stably higher BEC can be the matter. On the other hand, even if accuracy is lower than other methods, a method whose BEC is higher and BR is lower can be a choice when the improvements of partial important case are needed.}
    \end{description}

\section{Conclusion}
We compared methods which aim to improve degradation of classification result. The degradation aspect is important because the diversity of data samples in real use cases can cause misclassification and thus decreases the classification performance of important data samples.
Practitioners should be aware of trade-off between an accuracy and other non-accuracy metrics and dynamically choose each method depending on their purpose.

\section*{Acknowledgment}
    This work was partly supported by JST-Mirai Program Grant Number JPMJMI20B8, Japan.

\bibliographystyle{ACM-Reference-Format}
\bibliography{acmart}

\end{document}